\documentclass[conference]{IEEEtran}

\usepackage{cite}
\usepackage{amsmath,amssymb,amsfonts}
\usepackage{graphicx}
\usepackage{booktabs}
\usepackage{url}
\usepackage{xcolor}
\usepackage[hidelinks]{hyperref}
\usepackage{enumitem}
\usepackage{multirow}
\usepackage{pifont}
\usepackage{placeins}
\usepackage{array}
\usepackage{adjustbox}

\newcommand{\sd}[1]{\,{\fontsize{6pt}{7pt}\selectfont $\pm$#1}}

\renewcommand{\arraystretch}{0.92}

\begin{document}

\title{KPI-Aware Multimodal Anomaly Detection for 5G and Open RAN Telemetry}

\author{
\IEEEauthorblockN{Chiman Salavati\IEEEauthorrefmark{1}, Liang Wu\IEEEauthorrefmark{2}, Kelly Wan\IEEEauthorrefmark{2}, Mayank Darbari\IEEEauthorrefmark{2}, Liangjie Hong\IEEEauthorrefmark{2}}
\IEEEauthorblockA{\IEEEauthorrefmark{1}University of Connecticut, Storrs, Connecticut, USA; \IEEEauthorrefmark{2}Nokia, Sunnyvale, California, USA\\
chiman.salavati@uconn.edu, \{liang.wu, kelly.wan, mayank.darbari, liangjie.hong\}@nokia.com}
}

\maketitle

\begin{abstract}
Reliable anomaly detection is a deployment bottleneck for 5G/Open RAN observability and emerging self-optimizing 6G systems, where hundreds of heterogeneous KPIs, severe class imbalance, and strict alert budgets make purely numerical scores brittle.
We present KAC, a deployment-oriented KPI-aware multimodal anomaly-detection framework that uses KPI-level residual–text alignment to turn compact operator-style summaries into a learnable feedback signal over full-KPI temporal evidence.
This gives the detector broad numerical coverage like modern time-series models, while retaining the sparse KPI-specific reasoning that makes operator feedback sustainable in a future operator-facing workflow.
Across public telecom telemetry, enterprise-scale Open RAN observability data, and a Nokia benchmark built from real 5G RAN production packet traces with realism-hardened injected anomalies, KAC achieves strong threshold-based detection performance while remaining competitive on threshold-free ranking metrics. On TelecomTS, at a fixed 2.5\% false-positive-rate operating point, KAC achieves the highest recall among the evaluated residual-based methods while maintaining comparable precision; across the evaluated benchmarks, it remains robust under substantial class imbalance.
KAC is integrated in pre-production shadow mode with Nokia's GKE-based KPI construction pipeline, where it continuously scores telemetry windows alongside an internal packet-level anomaly detector for operational evaluation.
\end{abstract}


\section{Introduction}

Modern cellular observability pipelines collect heterogeneous telemetry from radio access networks (RANs), cloud/platform infrastructure, and packet-level measurements. In 5G and Open RAN, where RAN functions are increasingly disaggregated, virtualized, and connected through open interfaces, these telemetry streams include fine-grained key performance indicators (KPIs) such as radio quality, throughput, latency, resource usage, and packet-flow statistics~\cite{sun2024spotlight,feng2025telecomts}. These streams feed AIOps workflows for alerting, incident management, and diagnosis, yet noisy alert streams remain a persistent operational pain point~\cite{yu2024alertsurvey}. The challenge is amplified in Open RAN because anomaly detectors must jointly reason over radio-layer behavior and platform-level telemetry from disaggregated network functions~\cite{sun2024spotlight}. Recent telecom observability benchmarks further show that such data are heterogeneous, high-dimensional, and useful beyond forecasting, including anomaly detection, root-cause analysis, and multimodal reasoning~\cite{feng2025telecomts}. We target producing reliable anomaly scores from noisy, high-dimensional telecom telemetry windows under strict false-alarm constraints.

\begin{figure}[!t]
  \centering
  \includegraphics[width=\linewidth]{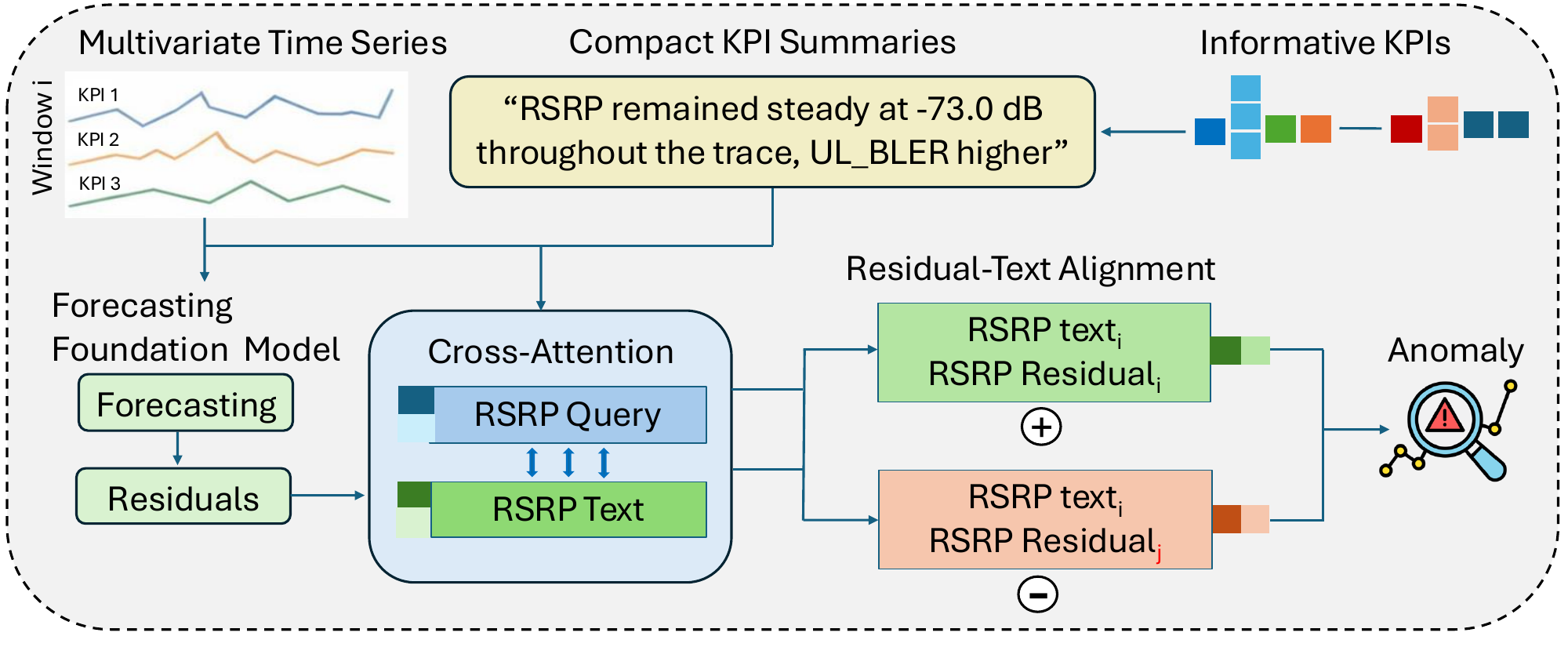}
    \caption{Overview of KAC. A frozen forecasting backbone yields per-KPI residual
    and uncertainty features, while a compact KPI-aware text summary is encoded into
    token representations. Learnable KPI queries extract per-KPI text representations,
    and a shared projector produces matched per-KPI residual representations; the two
    views of each KPI are aligned by an uncertainty-weighted contrastive objective,
    while gated multimodal fusion produces the window-level anomaly prediction.}
  \label{fig:overview}
\end{figure}

Multivariate time-series anomaly detection (MTSAD) has been studied extensively~\cite{schmidl2022evaluation,darban2024deep}, with deep models learning numerical representations of temporal behavior and inter-variable structure~\cite{su2019omnianomaly,audibert2020usad,tuli2022tranad,xu2022anomaly,deng2021gdn,zhao2020mtadgat}. However, these methods typically treat variables as numerical channels, graph nodes, or attention patterns rather than as named KPIs with operational meaning. In telecom observability, KPI identity and operating context often matter as much as the magnitude of a deviation. The same numerical change may be benign for one KPI yet indicate service degradation for another, and its interpretation can depend on the surrounding operating regime~\cite{feng2025telecomts}.

Time-series foundation models (TSFMs), including Chronos/Chronos-2~\cite{ansari2024chronos,ansari2025chronos2}, MOMENT~\cite{goswami2024moment}, Moirai~\cite{woo2024moirai}, and TimesFM~\cite{das2024timesfm}, provide strong pretrained temporal models and naturally suggest residual- or embedding-based detectors. Yet recent evidence shows that shallow forecasting-error or reconstruction-error reuse is often insufficient for anomaly detection~\cite{zhu2026oneliners}. In telecom settings, residual-only TSFM detectors are especially fragile in two cases. \textbf{Context extrapolation}: if an anomalous regime already begins inside the forecasting context, a strong forecaster may extrapolate it and produce small residuals even for an abnormal window. \textbf{Operating-context erasure}: residualization can suppress absolute KPI level; for example, a stable reference signal received power (RSRP) series may be forecast accurately both near the tower and at the cell edge, even though its absolute level is essential for interpreting block error rate (BLER), throughput, and packet-level indicators. These failure modes motivate combining full-KPI residual evidence with compact KPI-aware context rather than relying on forecast error alone.

We present \textbf{KAC}, a multimodal telecom anomaly detector that fuses two complementary modalities of each KPI window: a \emph{numerical} modality and a \emph{textual} modality. The numerical branch keeps full-KPI residual and uncertainty features from a frozen forecasting time-series foundation model, preserving broad temporal coverage across all KPIs. The textual branch encodes a compact operator-style summary that names the salient KPIs and their operating context under a token budget, reintroducing the KPI identity and absolute-level cues that residualization tends to erase. To make the two modalities reinforce each other rather than be concatenated as independent feature blocks, KAC uses \emph{contrastive learning}: learnable KPI queries extract KPI-specific text representations, a shared residual projector produces matched KPI-level residual representations, and a KPI-level contrastive objective aligns the text and residual views of each KPI in a shared embedding space, so that the two modalities of a given KPI window retrieve each other within a training batch. This alignment uses language as a compact contextual supervision signal rather than a standalone anomaly detector, allowing KPI-aware context to calibrate dense foundation-model residual evidence. Optional uncertainty weighting down-weights residual evidence with wide prediction intervals. At inference time, KAC operates entirely on cached residual and text features without online LLM inference.

This work is grounded in both reproducible public evaluation and industrial integration. We evaluate on public 5G telemetry, enterprise-scale Open RAN telemetry, and ProdTrace-SA, a Nokia collaboration benchmark built from real 5G RAN production packet captures with eight realism-hardened injected anomaly families. KAC is integrated with Nokia's GKE-based KPI-construction pipeline in \emph{pre-production shadow mode}, where it scores cached windows for offline evaluation. Code, processed public-benchmark artifacts, and evaluation scripts are released for replication.\footnote{Code and reproducibility artifact: \url{https://github.com/ChimanSalavati/kac-telecom-anomaly-detection}.} Our main contributions are:
\begin{itemize}[leftmargin=*,nosep]
    \item We introduce KAC, a telecom-oriented anomaly detector that combines full-KPI residual and uncertainty features from a frozen forecasting foundation model with compact operator-style context, addressing context extrapolation and operating-context erasure in residual-only detectors.

    \item We propose KPI-level residual--text alignment and uncertainty-aware contrastive learning that uses language as a compact contextual supervision signal for foundation-model residual representations.

    \item We develop a practical text pipeline for telecom observability, including leakage-aware summary construction, label-indicative masking, train-split-only saliency, and empirical leakage checks.

    \item We provide a comprehensive evaluation across public 5G telemetry, enterprise-scale Open RAN telemetry, and a Nokia collaboration benchmark built from real 5G RAN production packet captures with realism-hardened injected anomalies, together with low-FPR alert-budget analysis and pre-production shadow-integration evidence from Nokia's KPI-construction pipeline.
\end{itemize}

\section{Related Work}
\label{sec:related}

Telecom anomaly detection has moved from manual thresholding toward AI/ML-based monitoring for 5G and O-RAN environments~\cite{edozie2025telecomsurvey}. Recent systems emphasize reasoning across multiple telemetry layers; for example, SpotLight combines anomaly detection and localization over fine-grained RAN and platform metrics~\cite{sun2024spotlight}. In parallel, AIOps studies identify alert and incident management as a central failure-management task, highlighting the persistence of noisy alert streams in operational settings~\cite{yu2024alertsurvey}. These works motivate practical KPI anomaly detection, but they primarily operate on numerical telemetry and do not explicitly model KPI semantics or operator context.

Time-series foundation models such as Chronos/Chronos-2~\cite{ansari2024chronos,ansari2025chronos2}, MOMENT~\cite{goswami2024moment}, Moirai~\cite{woo2024moirai}, and TimesFM~\cite{das2024timesfm} provide pretrained forecasting and representation backbones for downstream time-series tasks. A common anomaly-detection strategy is to score reconstruction or forecasting errors, or to train shallow probes on frozen representations; however, recent evidence suggests that such direct reuse can be weak for anomaly detection~\cite{zhu2026oneliners}. Prompting-based LLM detectors, including SigLLM~\cite{alnegheimish2024sigllm} and recent systematic studies of LLMs for time-series anomaly detection~\cite{zhou2025llmts}, show promise on simple cases but remain less reliable than specialized detectors in subtle regimes, motivating the use of language as contextual supervision rather than as the detector itself.

Contrastive learning is widely used to improve MTSAD representations. CAROTS constructs causality-preserving and causality-disturbing augmentations~\cite{kim2025carots}; MPFormer uses multi-patch Transformer representations with contrastive learning~\cite{ma2024mpformer}; and DCdetector replaces reconstruction with a dual-attention contrastive criterion between in-patch and patch-wise views~\cite{yang2023dcdetector}. More broadly, supervised contrastive learning structures embeddings using label-aware positives and negatives~\cite{khosla2020supervised}, while CLIP demonstrates the effectiveness of aligning heterogeneous modalities in a shared representation space~\cite{radford2021clip}. Unlike prior MTSAD methods that contrast multiple numerical views of the same signal, KAC aligns KPI-level temporal and textual representations.

Multimodal time-series methods use textual or contextual supervision beyond purely numerical representations. The closest work, MindTS~\cite{hu2026mindts}, aligns time-series representations with fused exogenous and endogenous text and uses condensed textual representations for cross-modal reconstruction. In contrast, KAC aligns per-KPI forecasting-residual representations with budget-limited operator-style summaries, targeting KPI grounding and low-FPR telecom detection. Other related approaches use time--language alignment for understanding and classification~\cite{chen2025tsclip} and natural-language supervision for retrieval and channel-level grounding~\cite{chen2025trace}.

\section{Method}
\label{sec:method}

KAC addresses the context-extrapolation and operating-context-erasure failures of residual-only detection by feeding full-KPI residuals to the numerical branch and a fixed-budget salient-KPI summary to the text branch; KPI queries and a shared residual projector align the two views, with optional interval-width uncertainty weighting. Figure~\ref{fig:overview} provides an overview of the full pipeline.

\subsection{Problem Statement}
\label{sec:problem}

A KPI window $\mathbf{X} \in \mathbb{R}^{T \times K}$ contains $T$ timesteps and $K$ KPI channels, with entry $x_{t,k}$ (or $x_{i,t,k}$ for sample $i$) denoting the value of KPI $k$ at timestep $t$. Each window is paired with a natural-language summary $s$ that describes salient KPI behavior or operating context. The summary is not assumed to enumerate all $K$ KPIs: in high-dimensional telemetry, it may mention only a subset of informative indicators under a fixed token budget, while the numerical branch always receives the full multivariate window.

KAC learns a detector
\begin{equation}
    f_\theta(\mathbf{X}, s)=\hat{p}\in[0,1],
\end{equation}
where $\hat{p}$ is the predicted anomaly probability.

KAC decomposes as
\begin{equation}
    f_\theta(\mathbf{X}_i,s_i)
    =
    h_\theta\!\left(\Phi_C(\mathbf{X}_i), s_i\right),
\end{equation}
where $\Phi_C$ is a frozen forecasting foundation-model residual extractor with context length $C$, and $h_\theta$ is the trainable multimodal detector. In all reported experiments, $\Phi_C$ is instantiated with Chronos-2. The first $C$ timesteps serve as forecasting context, so the residual (forecast-target) sequence length is $T_r=T-C$.

Throughout the method section, $K$ denotes the number of KPI channels, $d$ the hidden dimension, $p$ the contrastive projection dimension, $B$ the batch size, and $L$ the maximum text length after tokenization.

\subsection{Foundation-Model Residual Feature Extraction}
\label{sec:chronos}

KAC first converts each KPI window into forecast-residual features
using a frozen forecasting time-series foundation model. We define
$\Phi_C$ as a residual-extraction interface with context length $C$:
given a KPI history, $\Phi_C$ returns point forecasts and, when
available, prediction intervals for the target timesteps. This interface is designed to support forecasting backbones that provide the residual and uncertainty information required by the numerical branch, although all experiments in this work use Chronos-2. We use Chronos-2 because its quantile forecasts support both residual construction and uncertainty weighting.

For sample $i$, KPI $k$, and residual timestep $\tau \in \{1,\ldots,T_r\}$,
the forecasting backbone produces quantile forecasts
$\hat{q}^{(0.05)}_{i,\tau,k}$,
$\hat{q}^{(0.50)}_{i,\tau,k}$, and
$\hat{q}^{(0.95)}_{i,\tau,k}$
for target value $x^{\star}_{i,\tau,k}=x_{i,\tau+C,k}$.
Suppressing the common indices $(i,\tau,k)$, let
$x^\star=x^\star_{i,\tau,k}$ and
$\hat{q}^{(\cdot)}=\hat{q}^{(\cdot)}_{i,\tau,k}$ for each quantile level $(\cdot)\in\{0.05,0.50,0.95\}$.
Using the same index-suppressed notation, we compute the residual feature vector
\begin{equation}
\boldsymbol{\rho}
=
[r,\; m^{-},\; m^{+},\; w,\; \tilde{r}],
\end{equation}
where
\begin{equation}
\begin{aligned}
r &= x^\star-\hat{q}^{(0.50)},\\
m^{-} &= x^\star-\hat{q}^{(0.05)},\\
m^{+} &= \hat{q}^{(0.95)}-x^\star,\\
w &= \hat{q}^{(0.95)}-\hat{q}^{(0.05)},\\
\tilde{r} &= \frac{|r|}{|x^\star|+\epsilon},
\end{aligned}
\end{equation}
and $\epsilon>0$ is a small constant added throughout for numerical stability. Restoring indices, we write $\boldsymbol{\rho}_{i,\tau,k}$ and denote its $a$-th component by $\rho^{(a)}_{i,\tau,k}$; $\rho^{(4)}_{i,\tau,k}=w$ is the raw prediction-interval width used for uncertainty weighting.

Residual features from all KPIs are stacked into $\mathbf{R}^{\mathrm{raw}}_i \in \mathbb{R}^{T_r \times 5K}$, where each KPI occupies a contiguous block of five channels:
\begin{equation}
R^{\mathrm{raw}}_{i,\tau,5(k-1)+a}
=
\rho^{(a)}_{i,\tau,k}.
\end{equation}

Before training, each residual channel is z-normalized using training-split statistics to produce the numerical input tensor $\mathbf{R}_i$. Raw prediction-interval widths are retained separately for uncertainty-weighted training.

\subsection{Compact Text Construction}
\label{sec:text_construction}

KAC uses text as compact semantic side information rather than as a replacement for numerical telemetry. Here a \emph{branch} refers to an input stream of the model: the numerical branch always receives the full residual tensor $\mathbf{R}_i \in \mathbb{R}^{T_r \times 5K}$, while the text branch receives a budget-limited summary $s_i$ that may mention only a subset of salient KPIs.

When summaries are not already provided, they are generated in two stages. First, a KPI selector is fit on the training split only and ranks channels by a saliency score. In label-free settings this score can use residual magnitude, residual variance, raw KPI variability, and operator-specified priority lists; when development labels are available, a supervised selector may be fit only on training labels. Category caps can be used so that a large family of similar counters does not consume the full text budget. Second, the top-ranked KPIs for each window are described with a fixed template or offline text generator using observed levels, trends, dispersion, and non-label timing or configuration context.

This construction handles large $K$ by compressing only the text side: one KPI is prioritized over another by the train-split saliency score and diversity constraints, but all unmentioned KPIs remain available to the numerical branch. The selector is fixed before validation/test/deployment and summaries are cached offline.

Before tokenization, we mask a fixed vocabulary $\mathcal{V}_{\mathrm{mask}}$ of label-indicative or diagnostic terms:
\begin{equation}
    s_i^{\mathrm{mask}}
    =
    \mathrm{Mask}(s_i; \mathcal{V}_{\mathrm{mask}}).
\end{equation}

Generated summaries never use anomaly labels or test statistics; saliency models are fit only on the training split.

For reproducibility, the artifact stores masked public-benchmark summaries, masking vocabularies, and the scripts/templates used to construct deterministic summaries. The masking vocabulary covers $30$ anomaly- or diagnostic-related terms (e.g.\ ``anomaly'', ``fault'', ``congestion'') that are replaced by a mask token before encoding. For restricted data, prompt templates and generation settings are documented for audit by authorized collaborators; because the underlying PCAP-derived windows cannot be redistributed, full public end-to-end reproduction is supported on TelecomTS and SpotLight.

\subsection{Text Encoding and KPI-Level Representations}
\label{sec:text_encoding}

The masked summary $s_i^{\mathrm{mask}}$ is tokenized to length $L$ and encoded with DistilBERT~\cite{sanh2019distilbert}; only the last two transformer blocks are unfrozen. We use DistilBERT for deployment-efficient text encoding. The resulting token states are projected to the KAC hidden dimension, $\mathbf{E}^{\mathrm{tok}}_i \in \mathbb{R}^{L \times d}$, and are used for both KPI-query attention and multimodal fusion.

KAC learns $K$ KPI query vectors, $\mathbf{Q}\in\mathbb{R}^{K \times d}$, one per KPI channel. These queries attend over $\mathbf{E}^{\mathrm{tok}}_i$ using masked multi-head attention~\cite{vaswani2017attention}, producing KPI-level text representations
$\mathbf{E}^{\mathrm{text}}_i \in \mathbb{R}^{K \times d}$.
Row $\mathbf{e}^{\mathrm{text}}_{i,k}$ corresponds to KPI $k$ and remains context-dependent even when the compact summary does not explicitly name that KPI.

For each KPI $k$, the corresponding five residual channels are mapped with a shared linear projection and mean-pooled over residual timesteps to produce $\mathbf{e}^{\mathrm{res}}_{i,k}\in\mathbb{R}^{d}$. Stacking over KPIs gives $\mathbf{E}^{\mathrm{res}}_i\in\mathbb{R}^{K \times d}$.

The text-derived and residual-derived KPI representations are projected and L2-normalized into a shared $p$-dimensional contrastive space:
$\mathbf{z}^{\mathrm{text}}_{i,k}, \mathbf{z}^{\mathrm{res}}_{i,k}\in\mathbb{R}^{p}$.
These paired embeddings are used in the uncertainty-weighted KPI contrastive loss in Section~\ref{sec:objective}.

\subsection{Gated Multimodal Fusion and Classification}
\label{sec:fusion}

The KPI-level representations support contrastive alignment, while the final anomaly score is produced from a fused sequence containing text tokens and the full-KPI residual sequence in the shared hidden space.

We first project the residual tensor to the hidden dimension,
$\mathbf{U}_i \in \mathbb{R}^{T_r \times d}$,
and compute a masked-mean text representation $\bar{\mathbf{e}}^{\mathrm{tok}}_i$ from $\mathbf{E}^{\mathrm{tok}}_i$.
A feature-wise gate
\begin{equation}
\mathbf{g}_i
=
\sigma(\mathrm{MLP}_{\mathrm{gate}}(\bar{\mathbf{e}}^{\mathrm{tok}}_i))
\in
\mathbb{R}^{d}
\end{equation}
modulates each residual timestep:
\begin{equation}
\tilde{\mathbf{U}}_{i,\tau}
=
\mathbf{U}_{i,\tau}
\odot
\mathbf{g}_i.
\end{equation}
The text summary therefore conditions which latent residual dimensions are emphasized before multimodal fusion.

The projected text sequence and gated residual sequence are concatenated along the sequence dimension and refined with a residual 1D convolution, dropout, and layer normalization. A learnable classification query then attends over the fused sequence to produce the window-level representation $\mathbf{h}_i \in \mathbb{R}^{d}$.

Finally, a two-layer MLP maps $\mathbf{h}_i$ to anomaly logit $\ell_i$ and predicted probability $\hat{p}_i=\sigma(\ell_i)$. In parallel, $\mathbf{h}_i$ is projected and L2-normalized to $\mathbf{z}^{\mathrm{sc}}_i \in \mathbb{R}^{p}$ for supervised contrastive learning.

\subsection{Uncertainty-Weighted Training Objective}
\label{sec:objective}

KAC is trained with three losses: binary cross-entropy for window-level anomaly detection, supervised contrastive learning over window embeddings, and uncertainty-weighted KPI-level contrastive learning between text-derived and residual-derived KPI representations.

Given ground-truth label $y_i \in \{0,1\}$ and anomaly logit $\ell_i$, we use class-weighted binary cross-entropy
\begin{equation}
\mathcal{L}_{\mathrm{BCE}}
=
-\frac{1}{B}
\sum_i
\left[
w_+ y_i \log \sigma(\ell_i)
+
(1-y_i)\log(1-\sigma(\ell_i))
\right],
\end{equation}
where $w_+=N^{-}/N^{+}$ is computed from training-set class counts.

Following supervised contrastive learning~\cite{khosla2020supervised}, we apply a contrastive loss to the window-level embeddings $\mathbf{z}^{\mathrm{sc}}_i$:
\begin{equation}
\mathcal{L}_{\mathrm{SupCon}}
=
-\frac{1}{|\mathcal{V}|}
\sum_{i\in\mathcal{V}}
\frac{1}{|\mathcal{P}(i)|}
\sum_{j\in\mathcal{P}(i)}
\log
\frac{
\exp(\mathbf{z}^{\mathrm{sc}}_i \cdot \mathbf{z}^{\mathrm{sc}}_j/\tau_{\mathrm{cl}})
}{
\sum_{m\neq i}
\exp(\mathbf{z}^{\mathrm{sc}}_i \cdot \mathbf{z}^{\mathrm{sc}}_m/\tau_{\mathrm{cl}})
},
\end{equation}
where $\mathcal{P}(i)$ denotes same-class positives for anchor $i$, $\mathcal{V}$ is the set of anchors with at least one positive in the batch, and $\tau_{\mathrm{cl}}=0.07$.

For each KPI $k$, we form a similarity matrix
\begin{equation}
\mathbf{S}_k
=
\mathbf{Z}^{\mathrm{text}}_k
(\mathbf{Z}^{\mathrm{res}}_k)^\top
/
\tau_{\mathrm{cl}},
\end{equation}
where $\mathbf{Z}^{\mathrm{text}}_k,\mathbf{Z}^{\mathrm{res}}_k \in \mathbb{R}^{B\times p}$ are the batch-stacked KPI projections. We then apply symmetric InfoNCE:
\begin{equation}
\ell^{\mathrm{KPI}}_k
=
\frac{1}{2}
\left[
\mathrm{CE}_{\mathrm{diag}}(\mathbf{S}_k)
+
\mathrm{CE}_{\mathrm{diag}}(\mathbf{S}_k^\top)
\right],
\end{equation}
where $\mathrm{CE}_{\mathrm{diag}}(\cdot)$ denotes the row-wise cross-entropy that treats each row's diagonal (matched-index) entry as the correct target, so that text- and residual-derived representations of the same KPI within the batch retrieve each other.

KPI-level contrastive learning should not treat every KPI equally: wide prediction intervals from the forecasting backbone indicate less reliable residual evidence. Using the raw interval-width feature $\rho^{(4)}_{i,\tau,k}$, we define
\begin{equation}
u_{i,k}
=
\frac{1}{T_r}
\sum_{\tau=1}^{T_r}
|\rho^{(4)}_{i,\tau,k}|,
\qquad
\tilde{\omega}_{i,k}
=
\frac{1}{u_{i,k}+\epsilon}.
\end{equation}
Weights are normalized within each sample and averaged across the batch:
\begin{equation}
\omega_k
=
\frac{1}{B}
\sum_{i=1}^{B}
\frac{
\tilde{\omega}_{i,k}
}{
K^{-1}\sum_{j=1}^{K}\tilde{\omega}_{i,j}+\epsilon
}.
\end{equation}
The uncertainty-weighted KPI contrastive loss is
\begin{equation}
\mathcal{L}_{\mathrm{KPI}}
=
\frac{
\sum_{k=1}^{K}
\omega_k \ell^{\mathrm{KPI}}_k
}{
\sum_{k=1}^{K}
\omega_k + \epsilon
}.
\end{equation}

The final training objective is
\begin{equation}
\mathcal{L}
=
\mathcal{L}_{\mathrm{BCE}}
+
\alpha \mathcal{L}_{\mathrm{SupCon}}
+
\beta \mathcal{L}_{\mathrm{KPI}},
\end{equation}
where $\alpha$ and $\beta$ control the contributions of the two contrastive terms.

\section{Experiments}
\label{sec:experiments}

\subsection{Experimental Setup}
\label{sec:setup}

We evaluate KAC on three complementary telecom anomaly-detection benchmarks that differ in dimensionality, telemetry source, and class imbalance (Table~\ref{tab:dataset_stats}). The evaluation compares KAC with trained residual classifiers, frozen time-series foundation-model probes, deep MTSAD baselines, classical one-class detectors, a paper-aligned Open RAN pipeline, and zero-shot LLM baselines. LLMs are reported separately because they receive no task-specific training or threshold tuning. All dataset splits used in our experiments are released in our public GitHub artifact for reproducibility, with the sole exception of ProdTrace-SA, which is proprietary to Nokia and cannot be redistributed.

\subsubsection{Datasets}
\label{sec:datasets}

\paragraph{TelecomTS~\cite{feng2025telecomts}}
TelecomTS is a public 5G observability benchmark with 16 KPIs sampled at 100\,ms. It contains mixed real jamming and synthetic telecom-rooted faults. We evaluate it under two scenarios, reported as \textbf{Scenario~1} and \textbf{Scenario~2} in Table~\ref{tab:telecomts_benchmark}: Scenario~1 uses the benchmark's balanced split, while Scenario~2 uses a severely imbalanced stress split with a natural anomaly rate of approximately $3.9\%$.

\paragraph{SpotLight~\cite{sun2024spotlight}}
SpotLight is a public enterprise Open RAN telemetry benchmark with 452 KPI channels and more than 100M telemetry points at 100\,ms resolution. It includes four anomaly families and six pairwise mixes, and its evaluation split is approximately class-balanced. For compact summaries, we rank KPIs with a train-only random-forest importance model on aggregated residual features from the chosen forecasting backbone, aggregate importances to KPI level with category-diversity constraints, and use deterministic templates to describe the selected KPI levels, trends, and dispersion.

\paragraph{ProdTrace-SA}
ProdTrace-SA is a Nokia collaboration benchmark built from real 5G RAN production packet captures from a major telecom equipment vendor, collected from operational network environments. Because naturally labeled attack incidents are unavailable, eight realism-hardened anomaly families are injected into the production traffic: \emph{DataExfiltration}, \emph{SignalingFlood}, \emph{PortScan}, \emph{SMPPSpam}, \emph{TCPAnomalies}, \emph{ExactPacketReplay}, \emph{SeqNumberManipulation}, and \emph{DuplicateBurst}. The injected anomalies preserve production-traffic characteristics through realism-hardening procedures such as RTT sampling, inter-arrival variation, and header/payload diversity. The same Nokia GKE-based KPI-construction infrastructure converts packet traces into $64$ KPI windows. For compact summaries, a label-free selector ranks packet-derived KPI channels from training windows using raw-window variability statistics, and the selected KPI statistics are summarized offline with an instruction-tuned LLM. Train, validation, and test windows are split by disjoint PCAPs to avoid capture-level leakage. ProdTrace-SA is Nokia-internal; public artifacts release the full evaluation pipeline and reproductions for public benchmarks. We use this production-traffic benchmark to evaluate sparse-positive behavior under a severe class-imbalance regime (anomaly rate $\approx\!15.7\%$) and to support the pre-production shadow-integration study in Section~\ref{sec:deployment_status}.

\begin{table}[t]
  \centering
  \caption{Dataset characteristics after KAC preprocessing. Raw scale reports the source telemetry before windowing.}
  \label{tab:dataset_stats}
  \small
  \setlength{\tabcolsep}{4pt}
  \begin{adjustbox}{max width=\columnwidth}
  \begin{tabular}{lccccc}
    \toprule
    Dataset & $K$ & $T$ & $C$ & $T_r$ & Raw scale \\
    \midrule
    TelecomTS  & 16  & 128 & 20 & 108 & 1.02M norm. / 120k anom. obs \\
    SpotLight  & 452 & 64  & 20 & 44  & 100M+ telemetry points \\
    ProdTrace-SA & 64  & 32  & 8  & 24  & 116 PCAPs, 22.6M packets \\
    \bottomrule
  \end{tabular}
  \end{adjustbox}
\end{table}

\subsubsection{Windowing and Labels}
\label{sec:windowing}

TelecomTS provides fixed windows. For SpotLight and ProdTrace-SA, point- or record-level labels are converted to window labels by marking a window anomalous if any constituent timestep or record is anomalous. SpotLight uses non-overlapping windows with $T=64$. ProdTrace-SA uses $T=32$ with stride $4$, resetting window construction at PCAP boundaries.

\subsubsection{Evaluation Protocol}
\label{sec:protocol}

We report threshold-free ranking metrics (AUROC and AP) and thresholded operating-point metrics (Precision, Recall, and F1). To ensure protocol fairness, all trainable methods use the same train/validation/test discipline. All trainable methods are fit on the training split only, with validation used exclusively for early stopping, model selection, and threshold calibration. Unsupervised methods similarly use validation only for threshold selection. Results are reported on the held-out test split across five seeds.

For thresholded Precision, Recall, and F1, we select a single decision threshold for each method and seed on the validation split by maximizing validation F1, and then apply that threshold unchanged to the test split. For operational alert-budget analysis, we additionally calibrate fixed-FPR thresholds on validation normal windows and evaluate them on the test split. Unless otherwise stated, fixed-budget results use a $2.5\%$ validation FPR target, corresponding to approximately $25$ false alerts per $1{,}000$ normal windows. Results are reported as mean$\pm$std across seeds; for ProdTrace-SA we additionally report bootstrap confidence intervals because of its small positive test count. In all result tables, \textbf{bold} marks the best value and \underline{underline} the second-best value within each metric column.

\subsubsection{Baselines}
\label{sec:baselines}

We compare against five baseline families. Residual and foundation-model baselines include Chronos-2+LR, QR-TAN, and frozen foundation-model embeddings from MOMENT~\cite{goswami2024moment}, TOTO~\cite{cohen2025toto}, and Mantis~\cite{feofanov2025mantis} with a trained linear classification head (+linear). Chronos-2+LR trains logistic regression on aggregated Chronos-2 residual features, while QR-TAN (ctx=20) is a residual-only temporal attention network using the same forecasting context as KAC.

End-to-end supervised baselines include PatchTST-classification~\cite{nie2023patchtst}, iTransformer~\cite{liu2024itransformer}, GPT4TS/OFA~\cite{zhou2023gpt4ts}, ModernTCN-classification~\cite{luo2024moderntcn}, and UniTS~\cite{gao2024units}. Deep MTSAD and one-class baselines include DCdetector~\cite{yang2023dcdetector}, TimesNet~\cite{wu2023timesnet}, ModernTCN~\cite{luo2024moderntcn}, MEMTO~\cite{song2023memto}, D3R~\cite{wang2023d3r}, Isolation Forest~\cite{liu2008isolation}, and LOF~\cite{breunig2000lof}. We also evaluate a simple relative-error rule that flags a window when the fraction of KPI-timestep cells with Chronos-2 relative forecast error $\tilde{r}=|r|/(|x^\star|+\epsilon)$ above $0.3$ exceeds a validation-tuned threshold. Finally, we include the SpotLight JVGAN--MRPI pipeline~\cite{sun2024spotlight} and zero-shot LLM baselines that receive KPI names and values without task-specific training or calibration. We do not include MindTS as a direct baseline because its formulation uses both exogenous and endogenous textual views, which are not consistently available across our telecom benchmarks; constructing additional exogenous text would change the information available to the compared methods.

\subsubsection{Implementation Details}
\label{sec:impl}

KAC uses hidden dimension $d=64$, contrastive dimension $p=128$, four attention heads, dropout $0.1$, maximum token length $L=128$, and contrastive temperature $\tau_{\mathrm{cl}}=0.07$. For the Chronos-2 instantiation used in all reported experiments, residuals and text summaries are cached before detector training. We train with AdamW~\cite{loshchilov2019adamw}, batch size $16$, weight decay $10^{-2}$, gradient clipping at $1.0$, and early stopping on validation F1. The unfrozen DistilBERT blocks use learning rate $2{\times}10^{-5}$; all remaining KAC parameters use $5{\times}10^{-4}$ on public benchmarks and $10^{-3}$ on ProdTrace-SA. For public benchmarks we use $(\alpha,\beta)=(1,1)$; for ProdTrace-SA we use $(0.1,0.3)$.

Code, preprocessing scripts, and evaluation configuration are released with the artifact.

\subsection{Results on TelecomTS}
\label{sec:res_telecomts}

Table~\ref{tab:telecomts_benchmark} shows that on the balanced TelecomTS split KAC attains the highest Precision ($0.961$), F1 ($0.960$), AUROC ($0.988$), and AP ($0.988$) among all supervised and unsupervised methods. The largest gains over residual-only and foundation-model baselines come from the compact text branch: in this relatively low-dimensional 16-KPI setting, operator-style summaries provide operating-context information that is difficult to recover from residuals alone.

Under the severely imbalanced stress split, KAC achieves the highest Precision ($0.900$) and ties for the highest F1 ($0.800$), while Mantis+linear attains the highest AUROC ($0.995$) and AP ($0.926$). This suggests that frozen foundation-model embeddings remain competitive for ranking under extremely small anomaly priors, whereas KAC achieves the highest precision and ties for the best F1 at the validation-selected operating point. Overall, the results support the central design motivation of KAC: combining full-KPI residual evidence with compact operating-context signals improves threshold-based anomaly detection beyond residual-only TSFM detectors and strong end-to-end supervised baselines.

\begin{table*}[t]
  \centering
  \caption{Experimental results on \textsc{TelecomTS} under balanced and imbalanced test settings.}
  \label{tab:telecomts_benchmark}
  \footnotesize
  \setlength{\tabcolsep}{3pt}
  \renewcommand{\arraystretch}{0.95}
  \begin{tabular*}{\textwidth}{@{\extracolsep{\fill}}lcccccccccc@{}}
    \toprule
    & \multicolumn{5}{c}{\textbf{Scenario 1}}
    & \multicolumn{5}{c}{\textbf{Scenario 2}} \\
    \cmidrule(lr){2-6} \cmidrule(lr){7-11}
    \textbf{Method}
    & \textbf{Prec.} & \textbf{Rec.} & \textbf{F1} & \textbf{AUROC} & \textbf{AP}
    & \textbf{Prec.} & \textbf{Rec.} & \textbf{F1} & \textbf{AUROC} & \textbf{AP} \\
    \midrule
    \multicolumn{11}{@{}l}{\emph{Supervised methods}} \\
    \addlinespace[1pt]
    QR-TAN (ctx=20)                              & \underline{.954}\sd{.025} & .894\sd{.018} & .923\sd{.007} & .974\sd{.007} & .978\sd{.005}
                                                 & \underline{.738}\sd{.092} & .700\sd{.000} & .716\sd{.042} & .919\sd{.026} & .762\sd{.035} \\
    Chronos-2 + LR                               & .916\sd{.000} & .887\sd{.000} & .901\sd{.000} & .967\sd{.000} & .970\sd{.000}
                                                 & .643\sd{.000} & \underline{.900}\sd{.000} & \underline{.750}\sd{.000} & .913\sd{.000} & \underline{.842}\sd{.000} \\
    MOMENT + linear     & .806\sd{.026} & .775\sd{.034} & .789\sd{.013} & .895\sd{.009} & .901\sd{.010}
                                                 & .132\sd{.022} & .620\sd{.130} & .218\sd{.038} & .796\sd{.016} & .245\sd{.027} \\
    Mantis + linear    & .931\sd{.000} & .927\sd{.000} & \underline{.929}\sd{.000} & \underline{.980}\sd{.000} & \underline{.982}\sd{.000}
                                                 & .667\sd{.000} & \textbf{1.00}\sd{.000} & \textbf{.800}\sd{.000} & \textbf{.995}\sd{.000} & \textbf{.926}\sd{.000} \\
    TOTO + linear           & .715\sd{.019} & .807\sd{.042} & .758\sd{.010} & .840\sd{.003} & .854\sd{.003}
                                                 & .250\sd{.000} & .200\sd{.000} & .222\sd{.000} & .591\sd{.021} & .224\sd{.029} \\
    PatchTST              & .872\sd{.020} & .846\sd{.017} & .859\sd{.009} & .930\sd{.005} & .934\sd{.007}
                                                 & .204\sd{.027} & .380\sd{.098} & .264\sd{.045} & .770\sd{.053} & .319\sd{.086} \\
    iTransformer      & .922\sd{.015} & .874\sd{.016} & .897\sd{.008} & .963\sd{.003} & .965\sd{.004}
                                                 & .447\sd{.044} & .520\sd{.040} & .478\sd{.020} & .861\sd{.022} & .618\sd{.017} \\
    GPT4TS                 & .837\sd{.027} & .158\sd{.026} & .264\sd{.037} & .780\sd{.015} & .769\sd{.012}
                                                 & .131\sd{.120} & .580\sd{.214} & .172\sd{.108} & .606\sd{.069} & .204\sd{.077} \\
    ModernTCN-cls        & .872\sd{.041} & .842\sd{.049} & .855\sd{.009} & .945\sd{.003} & .950\sd{.003}
                                                 & .595\sd{.094} & .360\sd{.080} & .435\sd{.033} & .897\sd{.023} & .574\sd{.026} \\
    UniTS                    & .946\sd{.027} & .911\sd{.018} & .928\sd{.008} & .974\sd{.005} & .978\sd{.005}
                                                 & .537\sd{.087} & .580\sd{.040} & .555\sd{.060} & .766\sd{.080} & .528\sd{.081} \\
    KAC (Ours)                                   & \textbf{.961}\sd{.018} & .960\sd{.010} & \textbf{.960}\sd{.004} & \textbf{.988}\sd{.002} & \textbf{.988}\sd{.004}
                                                 & \textbf{.900}\sd{.056} & .720\sd{.045} & \textbf{.800}\sd{.050} & \underline{.934}\sd{.038} & .839\sd{.035} \\
    \midrule
    \multicolumn{11}{@{}l}{\emph{Unsupervised and one-class methods}} \\
    \addlinespace[1pt]
    SpotLight   & .508\sd{.008} & \underline{.973}\sd{.012} & .667\sd{.004} & .581\sd{.022} & .560\sd{.015}
                                                 & .167\sd{.119} & .500\sd{.255} & .216\sd{.110} & .715\sd{.100} & .201\sd{.094} \\
    D3R                       & .599\sd{.060} & .862\sd{.122} & .700\sd{.016} & .668\sd{.033} & .641\sd{.018}
                                                 & .367\sd{.182} & .500\sd{.071} & .395\sd{.101} & .827\sd{.030} & .391\sd{.038} \\
    Relative-error rule                          & .518\sd{.000} & \textbf{.976}\sd{.000} & .677\sd{.000} & .651\sd{.000} & .651\sd{.000}
                                                 & .200\sd{.000} & .100\sd{.000} & .133\sd{.000} & .623\sd{.000} & .199\sd{.000} \\
    Isolation Forest                             & .608\sd{.085} & .775\sd{.125} & .670\sd{.014} & .677\sd{.021} & .710\sd{.019}
                                                 & .227\sd{.030} & .660\sd{.055} & .337\sd{.038} & .886\sd{.014} & .512\sd{.042} \\
    MEMTO                   & .731\sd{.034} & .608\sd{.063} & .662\sd{.035} & .721\sd{.017} & .729\sd{.019}
                                                 & .274\sd{.112} & .720\sd{.084} & .382\sd{.112} & .811\sd{.028} & .440\sd{.020} \\
    TimesNet               & .508\sd{.004} & .923\sd{.021} & .655\sd{.005} & .625\sd{.005} & .592\sd{.005}
                                                 & .109\sd{.034} & .200\sd{.122} & .130\sd{.035} & .719\sd{.016} & .147\sd{.035} \\
    ModernTCN            & .506\sd{.002} & .915\sd{.010} & .652\sd{.004} & .627\sd{.006} & .592\sd{.007}
                                                 & .090\sd{.008} & .380\sd{.045} & .146\sd{.014} & .714\sd{.004} & .131\sd{.007} \\
    DCdetector         & .517\sd{.013} & .816\sd{.095} & .631\sd{.025} & .537\sd{.015} & .536\sd{.016}
                                                 & .017\sd{.024} & .180\sd{.249} & .031\sd{.043} & .456\sd{.029} & .047\sd{.002} \\
    LOF                                          & .545\sd{.000} & .644\sd{.000} & .590\sd{.000} & .607\sd{.000} & .624\sd{.000}
                                                 & .053\sd{.000} & .400\sd{.000} & .094\sd{.000} & .600\sd{.000} & .068\sd{.000} \\
    \bottomrule
  \end{tabular*}
\end{table*}

Table~\ref{tab:alert_budget} evaluates detectors at a fixed low-FPR operating point on TelecomTS corresponding to approximately $25$ false alerts per $1{,}000$ normal windows. KAC achieves the highest recall at this alert budget while maintaining comparable precision. This operating regime is important in practice because telecom operators often tune anomaly detectors by alert volume rather than by full-curve averages.

We report fixed-budget analysis only on TelecomTS. SpotLight is class-balanced, so threshold-based F1 already reflects a meaningful operating point and AUROC/AP characterize performance across the full operating range. ProdTrace-SA's test split is only $\sim$15.7\% positive, making fixed-FPR recall estimates too unstable for reliable method ranking. We therefore rely on AUROC/AP for ProdTrace-SA.

\begin{table}[t]
  \centering
  \caption{Fixed alert-budget evaluation on \textsc{TelecomTS}. All methods are evaluated at $2.5\%$ FPR, corresponding to $25$ false alerts per $1{,}000$ normal windows.}
  \label{tab:alert_budget}
  \footnotesize
  \setlength{\tabcolsep}{4.5pt}
  \begin{adjustbox}{max width=\columnwidth}
  \begin{tabular}{@{}lcccc@{}}
    \toprule
    \textbf{Method} & \textbf{FPR budget} & \textbf{Recall} & \textbf{Prec.} & \textbf{False alerts} \\
                    &                     &                 &                & \textbf{/1k normal} \\
    \midrule
    QR-TAN (ctx=20)      & 2.5\% & .85 & .971 & 25 \\
    Chronos-2 + LR       & 2.5\% & .80 & .970 & 25 \\
    KAC (Ours)           & 2.5\% & \textbf{.93} & \textbf{.974} & 25 \\
    \bottomrule
  \end{tabular}
  \end{adjustbox}
\end{table}

Table~\ref{tab:llm_benchmark} reports zero-shot frontier LLMs operating directly on raw KPI matrices and KPI names. This is not intended as a fairness-matched comparison to a trained detector; rather, it evaluates the alternative design of delegating anomaly detection to an LLM without task-specific training, threshold calibration, or cached residual features. Exact prompts, model identifiers, evaluation dates, decoding parameters, parsing rules, and cached responses are released in the artifact for reproducibility. Across all benchmarks, zero-shot prompting remains substantially weaker than trained telecom-specific detectors, particularly under subtle or highly imbalanced regimes. These results support the KAC design choice of using language as a compact contextual side channel rather than relying on end-to-end LLM reasoning for anomaly detection.

\begin{table}[t]
  \centering
  \caption{Zero-shot LLM prompting versus trained KAC.}
  \label{tab:llm_benchmark}
  \footnotesize
  \setlength{\tabcolsep}{7pt}
  \begin{adjustbox}{max width=\columnwidth}
  \begin{tabular}{@{}llccc@{}}
    \toprule
    \textbf{Dataset} & \textbf{Model} & \textbf{F1} & \textbf{AUROC} & \textbf{AP} \\
    \midrule
    \multirow{4}{*}{TelecomTS}
      & Claude Opus 4.7 & .384 & .689 & .613 \\
      & GPT-5           & .607 & .651 & .612 \\
      & Gemini 2.5 Pro  & .453 & .523 & .543 \\
      & KAC (Ours)      & \textbf{.960} & \textbf{.988} & \textbf{.988} \\
    \midrule
    \multirow{4}{*}{ProdTrace-SA}
      & Claude Opus 4.7 & .250 & .627 & .521 \\
      & GPT-5           & .273 & .377 & .127 \\
      & Gemini 2.5 Pro  & .293 & .690 & .346 \\
      & KAC (Ours)      & \textbf{.742} & \textbf{.933} & \textbf{.876} \\
    \midrule
    \multirow{4}{*}{SpotLight}
      & Claude Opus 4.7 & .567 & .619 & .532 \\
      & GPT-5           & .641 & .692 & .616 \\
      & Gemini 2.5 Pro  & .602 & .660 & .639 \\
      & KAC (Ours)      & \textbf{.950} & \textbf{.986} & \textbf{.984} \\
    \bottomrule
  \end{tabular}
  \end{adjustbox}
\end{table}

SpotLight stresses high-dimensional Open RAN telemetry and strong residual structure~\cite{sun2024spotlight}. Table~\ref{tab:spotlight_benchmark} shows that KAC achieves the strongest F1 and Recall among all evaluated methods, leading the next strongest baseline (iTransformer) by $0.010$ F1.

The high-dimensional regime highlights a different behavior than TelecomTS. End-to-end-trained patch, channel-token, and convolutional classifiers remain highly competitive, but KAC's combination of full-KPI residual evidence and KPI-level multimodal alignment retains the strongest threshold-based detection performance. Unlike TelecomTS, the primary KAC gain on SpotLight comes from KPI-level contrastive alignment rather than from text fusion alone: the full residual tensor carries the dominant numerical evidence, while the text representation conditions residual fusion and provides the KPI-level alignment signal.

\begin{table}[t]
  \centering
  \caption{Experimental results on \textsc{SpotLight}.}
  \label{tab:spotlight_benchmark}
  \footnotesize
  \setlength{\tabcolsep}{2pt}
  \begin{adjustbox}{max width=\columnwidth}
  \begin{tabular}{@{}lccccc@{}}
    \toprule
    \textbf{Method} & \textbf{Prec.} & \textbf{Rec.} & \textbf{F1} & \textbf{AUROC} & \textbf{AP} \\
    \midrule
    \multicolumn{6}{l}{\emph{Supervised}} \\
    QR-TAN (ctx=20)                      & .927\sd{.005} & .945\sd{.024} & .936\sd{.011} & .978\sd{.009} & .978\sd{.007} \\
    Chronos-2 + LR                       & \underline{.944}\sd{.000} & .924\sd{.000} & .934\sd{.000} & .978\sd{.000} & .982\sd{.000} \\
    MOMENT + linear      & .922\sd{.014} & .847\sd{.049} & .882\sd{.021} & .965\sd{.003} & .957\sd{.005} \\
    Mantis + linear     & .923\sd{.000} & .945\sd{.000} & .934\sd{.000} & .967\sd{.000} & .973\sd{.000} \\
    TOTO + linear            & .831\sd{.012} & .812\sd{.008} & .821\sd{.004} & .913\sd{.001} & .915\sd{.001} \\
    PatchTST      & .934\sd{.008} & .944\sd{.030} & .939\sd{.015} & \textbf{.987}\sd{.004} & \textbf{.987}\sd{.003} \\
    iTransformer & .926\sd{.010} & \underline{.954}\sd{.017} & \underline{.940}\sd{.004} & .979\sd{.002} & .969\sd{.005} \\
    GPT4TS         & .845\sd{.039} & .657\sd{.047} & .737\sd{.027} & .871\sd{.018} & .856\sd{.025} \\
    ModernTCN-cls & \textbf{.945}\sd{.003} & .900\sd{.024} & .921\sd{.011} & .983\sd{.003} & .983\sd{.002} \\
    UniTS            & .924\sd{.015} & .918\sd{.029} & .920\sd{.009} & .979\sd{.002} & .977\sd{.003} \\
    KAC (Ours)                           & .939\sd{.006} & \textbf{.961}\sd{.012} & \textbf{.950}\sd{.004} & \underline{.986}\sd{.006} & \underline{.984}\sd{.006} \\
    \midrule
    \multicolumn{6}{l}{\emph{Unsupervised / one-class}} \\
    SpotLight pipeline & .658\sd{.010} & .910\sd{.018} & .763\sd{.009} & .773\sd{.017} & .720\sd{.037} \\
    ModernTCN    & .751\sd{.010} & .873\sd{.050} & .807\sd{.027} & .877\sd{.018} & .871\sd{.015} \\
    MEMTO           & .683\sd{.003} & .933\sd{.026} & .788\sd{.010} & .784\sd{.013} & .707\sd{.041} \\
    D3R               & .679\sd{.002} & .938\sd{.019} & .788\sd{.008} & .777\sd{.002} & .692\sd{.008} \\
    Relative-error rule                  & .714\sd{.000} & .866\sd{.000} & .783\sd{.000} & .779\sd{.000} & .708\sd{.000} \\
    TimesNet       & .741\sd{.015} & .820\sd{.007} & .779\sd{.006} & .858\sd{.011} & .849\sd{.019} \\
    Isolation Forest                     & .682\sd{.007} & .823\sd{.062} & .745\sd{.023} & .813\sd{.010} & .775\sd{.019} \\
    DCdetector & .726\sd{.061} & .671\sd{.230} & .673\sd{.091} & .804\sd{.005} & .758\sd{.011} \\
    LOF                          & .500\sd{.000} & .024\sd{.000} & .046\sd{.000} & .815\sd{.000} & .731\sd{.000} \\
    \bottomrule
  \end{tabular}
  \end{adjustbox}
\end{table}

\subsection{Results on ProdTrace-SA}
\label{sec:res_production}

ProdTrace-SA evaluates detector behavior under severe class imbalance using its packet-derived KPI traces processed through Nokia's KPI construction pipeline. Table~\ref{tab:production_benchmark} shows that KAC achieves the strongest mean F1 ($0.742$), AP ($0.876$), and tied-best AUROC ($0.933$) among the evaluated methods. These results indicate that KPI-level multimodal alignment remains effective under sparse-positive conditions representative of pre-production telecom monitoring. Although confidence intervals remain wide because of the sparse-positive regime, KAC avoids the collapse and threshold-transfer failures observed in several supervised and one-class baselines while maintaining stable threshold-based performance.

\begin{table}[!htbp]
  \centering
  \caption{Experimental results on ProdTrace-SA.}
  \label{tab:production_benchmark}
  \small
  \setlength{\tabcolsep}{6pt}
  \begin{adjustbox}{max width=\columnwidth}
  \begin{tabular}{@{}lccc@{}}
    \toprule
    \textbf{Method} & \textbf{F1} & \textbf{AUROC} & \textbf{AP} \\
    \midrule
    \multicolumn{4}{l}{\emph{Supervised}} \\
    QR-TAN (ctx=20)                      & \underline{.737}\sd{.078} & .905\sd{.056} & .815\sd{.085} \\
    Chronos-2 + LR                       & .500\sd{.000} & .637\sd{.000} & .518\sd{.000} \\
    Residual+Level (LR)                  & .378\sd{.000} & .669\sd{.000} & .469\sd{.000} \\
    TOTO + linear           & .047\sd{.079}  & .409\sd{.064}  & .135\sd{.014} \\
    PatchTST      & .502\sd{.080} & .876\sd{.064} & .821\sd{.094} \\
    iTransformer & .411\sd{.081} & \textbf{.933}\sd{.016} & \underline{.870}\sd{.018} \\
    GPT4TS         & .214\sd{.141} & .451\sd{.146} & .387\sd{.138} \\
    KAC (Ours)                           & \textbf{.742}\sd{.077} & \textbf{.933}\sd{.048} & \textbf{.876}\sd{.066} \\
    \midrule
    \multicolumn{4}{l}{\emph{Unsupervised / one-class}} \\
    Relative-error rule                  & .615\sd{.000} & .797\sd{.000} & .572\sd{.000} \\
    MEMTO           & .540\sd{.097} & \underline{.918}\sd{.003} & .708\sd{.056} \\
    TimesNet       & .479\sd{.064} & .851\sd{.056} & .580\sd{.049} \\
    ModernTCN    & .459\sd{.030} & .914\sd{.005} & .700\sd{.027} \\
    D3R               & .455\sd{.076} & .823\sd{.131} & .666\sd{.074} \\
    LOF                                  & .438\sd{.000} & .875\sd{.000} & .683\sd{.000} \\
    Isolation Forest                     & .276\sd{.030} & .593\sd{.122} & .293\sd{.126} \\
    DCdetector & .222\sd{.037} & .490\sd{.080} & .176\sd{.046} \\
    SpotLight pipeline~\cite{sun2024spotlight} & .265\sd{.026} & .511\sd{.049} & .207\sd{.055} \\
    \bottomrule
  \end{tabular}
  \end{adjustbox}
\end{table}

\subsection{Paired Significance}
\label{sec:significance}

To avoid over-interpreting small F1 gaps under sparse positives, we report paired KAC$-$baseline $\Delta$F1 with $95\%$ bootstrap confidence intervals over five seeds (Table~\ref{tab:significance}); positive values favor KAC, and a CI that excludes zero indicates a seed-consistent advantage. We do not run formal significance tests because five paired seeds are underpowered. KAC's threshold-F1 lead is consistent and CI-separated on TelecomTS against all four baselines, and is large and CI-separated on ProdTrace-SA against iTransformer, PatchTST, and Mantis.

The only case where the CI includes zero is KAC versus QR-TAN on ProdTrace-SA. Although KAC remains ahead in mean F1 ($+0.028$), the bootstrap CI $[-0.080,\,0.114]$ crosses zero, indicating that the observed difference is not reliably separable from seed-level variation in this sparse-positive setting. This result is consistent with QR-TAN already operating on the same Chronos-2 residual evidence used by KAC.

\begin{table}[t]
  \centering
  \caption{Paired KAC$-$baseline $\Delta$F1 (mean and $95\%$ bootstrap CI over five seeds; positive favours KAC).}
  \label{tab:significance}
  \small
  \setlength{\tabcolsep}{6pt}
  \begin{adjustbox}{max width=\columnwidth}
  \begin{tabular}{@{}lcc@{}}
    \toprule
    \textbf{Baseline} & \textbf{$\Delta$F1 TelecomTS} & \textbf{$\Delta$F1 ProdTrace-SA} \\
    \midrule
    QR-TAN       & $+.034$ [.024,\,.042] & $+.028$ [$-$.080,\,.114] \\
    iTransformer & $+.060$ [.050,\,.069] & $+.355$ [.309,\,.401] \\
    PatchTST     & $+.098$ [.089,\,.105] & $+.263$ [.221,\,.305] \\
    Mantis       & $+.027$ [.020,\,.035] & $+.229$ [.190,\,.268] \\
    \bottomrule
  \end{tabular}
  \end{adjustbox}
\end{table}

\subsection{Component Ablation}
\label{sec:ablation}

Table~\ref{tab:ablation} isolates the contribution of the compact text branch and KPI-level contrastive alignment. V1 uses residual-only gated fusion, V2 adds text fusion, and V3 corresponds to the full KAC model with KPI-level alignment. On TelecomTS, the largest gain comes from adding the text branch, indicating that operator-style summaries recover operating-context information that is weakly expressed in residuals alone. On SpotLight, text fusion alone provides only a modest gain, while the larger V2-to-V3 improvement shows that KPI-level alignment is the primary multimodal contribution in high-dimensional telemetry.

\begin{table*}[t]
  \centering
  \caption{Component ablation on benchmarks. V1 removes text, V2 adds text fusion, and V3 adds KPI-level contrastive alignment.}
  \label{tab:ablation}
  \footnotesize
  \setlength{\tabcolsep}{3pt}
  \renewcommand{\arraystretch}{0.95}
  \scalebox{0.95}{
  \begin{tabular*}{\textwidth}{@{\extracolsep{\fill}}llccccc@{}}
    \toprule
    \textbf{Dataset} & \textbf{Variant} & \textbf{Precision} & \textbf{Recall} & \textbf{F1} & \textbf{AUROC} & \textbf{AP} \\
    \midrule
    \multirow{3}{*}{\textsc{TelecomTS}} & V1: Residual-only & 0.943\sd{0.015} & 0.895\sd{0.016} & 0.918\sd{0.010} & 0.972\sd{0.004} & 0.976\sd{0.002} \\
     & V2: + Text fusion & \textbf{0.967}\sd{0.010} & 0.953\sd{0.011} & \textbf{0.960}\sd{0.007} & 0.987\sd{0.004} & \textbf{0.988}\sd{0.005} \\
     & V3: KAC (ours) & 0.961\sd{0.016} & \textbf{0.960}\sd{0.009} & \textbf{0.960}\sd{0.004} & \textbf{0.988}\sd{0.002} & \textbf{0.988}\sd{0.003} \\
    \addlinespace[1pt]
    \midrule
    \addlinespace[1pt]
    \multirow{3}{*}{\textsc{SpotLight}} & V1: Residual-only & 0.920\sd{0.008} & 0.953\sd{0.013} & 0.937\sd{0.007} & 0.975\sd{0.006} & 0.968\sd{0.009} \\
     & V2: + Text fusion & 0.931\sd{0.012} & 0.951\sd{0.014} & 0.940\sd{0.010} & 0.974\sd{0.011} & 0.967\sd{0.022} \\
     & V3: KAC (ours) & \textbf{0.939}\sd{0.006} & \textbf{0.961}\sd{0.011} & \textbf{0.950}\sd{0.003} & \textbf{0.986}\sd{0.005} & \textbf{0.984}\sd{0.006} \\
    \bottomrule
  \end{tabular*}}
\end{table*}

A natural concern is that KAC's gains arise from reintroducing absolute KPI levels removed by residualization rather than from language alignment. To test this, we add a non-language baseline (Residual+Level) that augments Chronos residual aggregates with raw per-KPI window statistics and train-split saliency features, without a text branch. Although adding absolute levels improves over residual-only detection, it remains below KAC on all three benchmarks and falls substantially behind on ProdTrace-SA (0.378 vs.\ 0.742 F1; Table~\ref{tab:production_benchmark}). These results indicate that KAC's gains cannot be explained solely by recovering absolute KPI levels.

Uncertainty weighting contributes less than text fusion or KPI-level alignment on the public benchmarks, but consistently improves stability under high-variance conditions. We therefore use it primarily as a robustness mechanism in the ProdTrace-SA evaluation.

\subsection{Text Leakage and Context Recovery Analysis}
\label{sec:leakage_analysis}

TelecomTS descriptions originate from anomaly-conditioned prompts, so we evaluate whether the text branch exploits label shortcuts rather than learning meaningful KPI context. A keyword audit finds no explicit anomaly-type tokens in the descriptions. Moreover, the performance benefit of text survives aggressive masking of label-indicative patterns, while label-independent KPI-derived summaries constructed from raw KPI statistics still improve over the residual-only baseline. These results suggest that the text branch contributes information beyond explicit anomaly labels.

We further evaluate text--residual alignment through contradictory-text controls. Replacing anomaly descriptions with normal descriptions substantially reduces F1, while increasing the text budget from $128$ to $512$ tokens degrades performance and makes the model largely insensitive to text permutations. Together, these results are consistent with the text branch acting mainly as a contextual calibration signal that suppresses false positives under different operating regimes, rather than relying primarily on explicit label leakage.

\section{Pre-Production Shadow Integration and Operational Cost}
\label{sec:deployment}

KAC is designed for low-cost pre-production shadow integration in telecom KPI pipelines. Foundation-model residual extraction, instantiated with Chronos-2 in the current shadow pipeline, and summary generation are performed offline or cached; the shadow scorer does not make online LLM calls. The latency in Table~\ref{tab:deployment_cost} measures the KAC scoring path on cached residual/summary inputs, including DistilBERT text encoding and lightweight fusion/classification heads, but excluding offline Chronos-2 extraction and summary generation. Measurements are reported for CPU execution, reflecting the current shadow-scoring configuration and demonstrating that the online KAC scorer does not require GPU allocation. The KAC-specific modules account for only $\sim$$0.21\%$ of parameters beyond the DistilBERT backbone. Single-window scoring is below the $100$\,ms telemetry cadence used in the evaluated benchmarks (batch-1 median $14.7$\,ms, p95 $17.7$\,ms), while batching improves throughput at the cost of higher batch latency. In the shadow configuration, Chronos-2 extraction is parallelized across KPI streams, and text-summary refresh is event-triggered rather than per-window, amortizing language-processing cost over operational intervals.

\begin{table}[t]
  \centering
  \caption{Operational cost of the KAC pre-production shadow scorer. Left: parameter breakdown. Right: CPU execution latency on cached residual/summary inputs}
  \label{tab:deployment_cost}
  \small
  \setlength{\tabcolsep}{3pt}
  \begin{minipage}[t]{0.48\linewidth}
    \centering
    \begin{adjustbox}{max width=\linewidth}
    \begin{tabular}{@{}lr@{}}
      \toprule
      \textbf{Component} & \textbf{Params} \\
      \midrule
      DistilBERT, frozen   & 52.19\,M (78.5\%) \\
      DistilBERT, last 2   & 14.18\,M (21.3\%) \\
      KAC fusion + heads   & 0.14\,M (0.21\%) \\
      Learnable queries    & 1{,}088 (0.00\%) \\
      \midrule
      \textbf{Total}       & \textbf{66.51\,M} \\
      \emph{Trainable}     & \emph{14.32\,M (21.5\%)} \\
      \bottomrule
    \end{tabular}
    \end{adjustbox}
  \end{minipage}%
  \hfill
  \begin{minipage}[t]{0.48\linewidth}
    \centering
    \begin{adjustbox}{scale=0.88}
    \begin{tabular}{@{}rrrr@{}}
      \toprule
      \textbf{Batch} & \textbf{Med.} & \textbf{p95} & \textbf{Tput.} \\
                     & \textbf{(ms)} & \textbf{(ms)} & \textbf{(w/s)} \\
      \midrule
      1  & 14.7  & 17.7  & 68.0  \\
      16 & 156.8 & 158.5 & 102.0 \\
      64 & 572.4 & 608.0 & 111.8 \\
      \bottomrule
    \end{tabular}
    \end{adjustbox}
  \end{minipage}
\end{table}

\subsection{Shadow Integration Status}
\label{sec:deployment_status}

KAC is integrated in pre-production shadow mode with Nokia's GKE-based KPI construction pipeline, which builds per-window KPI tensors and cached operator-style summaries from 5G RAN captures. The infrastructure uses Kubernetes-based telemetry services including Kong, NATS, and MinIO for ingestion, streaming, and object storage.

In the current shadow configuration, KAC runs alongside Nokia's internal packet-level anomaly detector. The pipeline passes KPI windows to KAC, cached residuals and summaries are loaded, and the resulting anomaly scores are logged for offline comparison against the internal detector. KAC does not currently promote incidents, trigger operator-facing alerts, or initiate mitigation.

The pre-production evidence reported in this paper focuses on four operational aspects: detection quality across heterogeneous telecom benchmarks; behavior under fixed low-FPR alert budgets; robustness under severe class imbalance; and operational cost, including CPU execution latency and parameter footprint (Table~\ref{tab:deployment_cost}). Together, these results support deployment compatibility and stable pre-production behavior without online LLM inference. We deliberately do \emph{not} claim operator-facing impact: detection delay, alert-volume reduction, incident-level precision/recall, and head-to-head comparison against Nokia's internal packet-level anomaly detector require a controlled operator-facing rollout and are left to future deployment studies (Section~\ref{sec:limitations}).

\section{Limitations and Future Work}
\label{sec:limitations}

KAC achieves strong threshold-based detection and favorable fixed low-FPR behavior on TelecomTS, but it does not dominate every metric. Mantis+linear remains stronger on AUROC/AP in the imbalanced TelecomTS stress split, and PatchTST remains strongest on AUROC/AP for SpotLight, showing that frozen or end-to-end numerical baselines can remain competitive on ranking metrics. ProdTrace-SA's anomaly labels are generated through realism-hardened injection of multiple anomaly families rather than naturally occurring operator-labeled incidents, and the sparse-positive regime yields wide confidence intervals. Because ProdTrace-SA and the shadow pipeline are Nokia-internal, they provide supportive industrial evidence rather than independent public production validation; TelecomTS and SpotLight provide the fully reproducible public evidence base. Finally, the Nokia integration is limited to pre-production shadow mode, and the text modality has not yet been validated on noisy, naturally authored operator notes or missing-text conditions. Future work will evaluate KAC on naturally occurring operational incidents, broader fault scenarios, robustness to real operator feedback, and operator-facing deployments.

\section{Conclusion}

KAC is a telecom anomaly detector that uses KPI-level residual--text alignment to inject compact KPI-aware context into foundation-model residual representations, without online LLM inference. Across public 5G telemetry, enterprise-scale Open RAN observability data, and production-traffic-derived KPI benchmarks, KAC achieves strong threshold-based detection performance while remaining competitive on ranking metrics; on TelecomTS, it also improves recall under a fixed low-FPR alert budget. KAC is integrated in pre-production shadow mode with Nokia's GKE-based KPI-construction pipeline. These results support KPI-aware multimodal alignment as a practical direction for telecom observability.

\section*{GenAI Usage Disclosure}
We distinguish three uses of generative AI in this work. (i) \emph{Manuscript preparation}: generative AI was used only for grammar checking and sentence clarification of author-written text. (ii) \emph{Offline dataset text construction}: for ProdTrace-SA, the compact operator-style window summaries used by KAC's text branch were generated offline with an instruction-tuned LLM. (iii) \emph{Experimental baselines}: the zero-shot results use LLM APIs (Claude, GPT, Gemini) purely as comparison baselines, with no task-specific training.

\bibliographystyle{IEEEtran}
\bibliography{references}

@inproceedings{sun2024spotlight,
  author    = {Sun, Chuanhao and Pawar, Ujjwal and Khoja, Molham and Foukas, Xenofon and Marina, Mahesh K. and Radunovic, Bozidar},
  title     = {{SpotLight}: Accurate, Explainable and Efficient Anomaly Detection for {Open RAN}},
  booktitle = {Proc.\ ACM International Conference on Mobile Computing and Networking (MobiCom)},
  year      = {2024},
  doi       = {10.1145/3636534.3649380},
}

@article{feng2025telecomts,
  author        = {Feng, Austin and Varvarigos, Andreas and Panitsas, Ioannis and Fernandez, Daniela and Wei, Jinbiao and Guo, Yuwei and Chen, Jialin and Maatouk, Ali and Tassiulas, Leandros and Ying, Rex},
  title         = {{TelecomTS}: A Multi-Modal Observability Dataset for Time Series and Language Analysis},
  journal       = {arXiv preprint arXiv:2510.06063},
  year          = {2025},
  archivePrefix = {arXiv},
  eprint        = {2510.06063},
  doi           = {10.48550/arXiv.2510.06063}
}

@article{yu2024alertsurvey,
  author  = {Yu, Qingyang and Zhao, Nengwen and Li, Mingjie and Li, Zeyan and Wang, Honglin and Zhang, Wenchi and Sui, Kaixin and Pei, Dan},
  title   = {A Survey on Intelligent Management of Alerts and Incidents in {IT} Services},
  journal = {Journal of Network and Computer Applications},
  volume  = {224},
  pages   = {103842},
  year    = {2024},
  doi     = {10.1016/j.jnca.2024.103842}
}

@article{schmidl2022evaluation,
  author  = {Schmidl, Sebastian and Wenig, Phillip and Papenbrock, Thorsten},
  title   = {Anomaly Detection in Time Series: A Comprehensive Evaluation},
  journal = {Proceedings of the VLDB Endowment},
  volume  = {15},
  number  = {9},
  pages   = {1779--1797},
  year    = {2022},
  doi     = {10.14778/3538598.3538602}
}

@article{darban2024deep,
  author    = {Zamanzadeh Darban, Zahra and Webb, Geoffrey I. and Pan, Shirui and Aggarwal, Charu C. and Salehi, Mahsa},
  title     = {Deep Learning for Time Series Anomaly Detection: A Survey},
  journal   = {ACM Computing Surveys},
  volume    = {57},
  number    = {1},
  articleno = {15},
  numpages  = {42},
  year      = {2024},
  doi       = {10.1145/3691338}
}

@article{ansari2024chronos,
  author  = {Ansari, Abdul Fatir and Stella, Lorenzo and Turkmen, Caner and Zhang, Xiyuan and Mercado, Pedro and Shen, Huibin and Shchur, Oleksandr and Rangapuram, Syama Sundar and Pineda Arango, Sebastian and Kapoor, Shubham and Zschiegner, Jasper and Maddix, Danielle C. and Wang, Hao and Mahoney, Michael W. and Torkkola, Kari and Wilson, Andrew Gordon and Bohlke-Schneider, Michael and Wang, Yuyang},
  title   = {Chronos: Learning the Language of Time Series},
  journal = {Transactions on Machine Learning Research},
  year    = {2024},
  url     = {https://openreview.net/forum?id=gerNCVqqtR}
}

@article{ansari2025chronos2,
  author        = {Ansari, Abdul Fatir and Shchur, Oleksandr and K{\"u}ken, Jaris and Auer, Andreas and Han, Boran and Mercado, Pedro and Rangapuram, Syama Sundar and Shen, Huibin and Stella, Lorenzo and Zhang, Xiyuan and Goswami, Mononito and Kapoor, Shubham and Maddix, Danielle C. and Guerron, Pablo and Hu, Tony and Yin, Junming and Erickson, Nick and Desai, Prateek Mutalik and Wang, Hao and Rangwala, Huzefa and Karypis, George and Wang, Yuyang and Bohlke-Schneider, Michael},
  title         = {{Chronos-2}: From Univariate to Universal Forecasting},
  journal       = {arXiv preprint arXiv:2510.15821},
  year          = {2025},
  archivePrefix = {arXiv},
  eprint        = {2510.15821},
  doi           = {10.48550/arXiv.2510.15821}
}

@inproceedings{goswami2024moment,
  author    = {Goswami, Mononito and Szafer, Konrad and Choudhry, Arjun and Cai, Yifu and Li, Shuo and Dubrawski, Artur},
  title     = {{MOMENT}: A Family of Open Time-Series Foundation Models},
  booktitle = {Proceedings of the 41st International Conference on Machine Learning},
  series    = {Proceedings of Machine Learning Research},
  volume    = {235},
  year      = {2024},
  publisher = {PMLR}
}

@inproceedings{woo2024moirai,
  author    = {Woo, Gerald and Liu, Chenghao and Kumar, Akshat and Xiong, Caiming and Savarese, Silvio and Sahoo, Doyen},
  title     = {Unified Training of Universal Time Series Forecasting Transformers},
  booktitle = {Proceedings of the 41st International Conference on Machine Learning},
  series    = {Proceedings of Machine Learning Research},
  volume    = {235},
  year      = {2024},
  publisher = {PMLR}
}

@inproceedings{das2024timesfm,
  author    = {Das, Abhimanyu and Kong, Weihao and Sen, Rajat and Zhou, Yichen},
  title     = {A Decoder-Only Foundation Model for Time-Series Forecasting},
  booktitle = {Proceedings of the 41st International Conference on Machine Learning},
  series    = {Proceedings of Machine Learning Research},
  volume    = {235},
  pages     = {10148--10167},
  year      = {2024},
  publisher = {PMLR}
}

@inproceedings{gao2024units,
  author    = {Gao, Shanghua and Koker, Teddy and Queen, Owen and Hartvigsen, Thomas and Tsiligkaridis, Theodoros and Zitnik, Marinka},
  title     = {{UniTS}: A Unified Multi-Task Time Series Model},
  booktitle = {Advances in Neural Information Processing Systems},
  volume    = {37},
  year      = {2024}
}

@inproceedings{alnegheimish2024sigllm,
  author    = {Alnegheimish, Sarah and Nguyen, Linh and Berti-{\'E}quille, Laure and Veeramachaneni, Kalyan},
  title     = {Large Language Models Can Be Zero-Shot Anomaly Detectors for Time Series?},
  booktitle = {2024 IEEE International Conference on Data Science and Advanced Analytics},
  year      = {2024},
  publisher = {IEEE},
  archivePrefix = {arXiv},
  eprint    = {2405.14755}
}

@inproceedings{zhou2025llmts,
  author    = {Zhou, Zihao and Yu, Rose},
  title     = {Can {LLMs} Understand Time Series Anomalies?},
  booktitle = {International Conference on Learning Representations},
  year      = {2025}
}

@inproceedings{chen2025tsclip,
  author    = {Chen, Ziwen and Zhang, Xiaoyuan and Zhu, Ming},
  title     = {{TS}-{CLIP}: Time Series Understanding by {CLIP}},
  booktitle = {Proceedings of the 2025 Conference on Empirical Methods in Natural Language Processing},
  pages     = {4646--4664},
  year      = {2025},
  address   = {Suzhou, China},
  publisher = {Association for Computational Linguistics},
  doi       = {10.18653/v1/2025.emnlp-main.231}
}

@inproceedings{hu2026mindts,
  author    = {Hu, Shiyan and Jin, Jianxin and Shu, Yang and Chen, Peng and Yang, Bin and Guo, Chenjuan},
  title     = {Towards Multimodal Time Series Anomaly Detection with Semantic Alignment and Condensed Interaction},
  booktitle = {International Conference on Learning Representations},
  year      = {2026},
  archivePrefix = {arXiv},
  eprint        = {2603.21612},
  doi           = {10.48550/arXiv.2603.21612}
}

@inproceedings{chen2025trace,
  author    = {Chen, Jialin and Zhao, Ziyu and Nurbek, Gaukhar and Feng, Aosong and Maatouk, Ali and Tassiulas, Leandros and Gao, Yifeng and Ying, Rex},
  title     = {{TRACE}: Grounding Time Series in Context for Multimodal Embedding and Retrieval},
  booktitle = {Advances in Neural Information Processing Systems},
  year      = {2025},
  archivePrefix = {arXiv},
  eprint        = {2506.09114},
  doi           = {10.48550/arXiv.2506.09114}
}

@article{edozie2025telecomsurvey,
  author  = {Edozie, Enerst and Shuaibu, Aliyu Nuhu and Sadiq, Bashir Olaniyi and John, Ukagwu Kelechi},
  title   = {Artificial Intelligence Advances in Anomaly Detection for Telecom Networks},
  journal = {Artificial Intelligence Review},
  volume  = {58},
  pages   = {100},
  year    = {2025},
  doi     = {10.1007/s10462-025-11108-x}
}

@inproceedings{zhu2026oneliners,
  author    = {Zhu, Xiaokun and Carpentier, Louis and Verbeke, Mathias},
  title     = {When Foundation Models are One-Liners: Limitations and Future Directions for Time Series Anomaly Detection},
  booktitle = {International Conference on Learning Representations},
  year      = {2026},
  url       = {https://openreview.net/forum?id=H27kvyG4qf}
}

@inproceedings{zhao2020mtadgat,
  author    = {Zhao, Hang and Wang, Yujing and Duan, Juanyong and Huang, Congrui and Cao, Defu and Tong, Yunhai and Xu, Bixiong and Bai, Jing and Tong, Jie and Zhang, Qi},
  title     = {Multivariate Time-Series Anomaly Detection via Graph Attention Network},
  booktitle = {2020 IEEE International Conference on Data Mining},
  pages     = {841--850},
  year      = {2020},
  publisher = {IEEE},
  doi       = {10.1109/ICDM50108.2020.00093}
}

@inproceedings{audibert2020usad,
  author    = {Audibert, Julien and Michiardi, Pietro and Guyard, Fr{\'e}d{\'e}ric and Marti, S{\'e}bastien and Zuluaga, Maria A.},
  title     = {{USAD}: {UnSupervised} Anomaly Detection on Multivariate Time Series},
  booktitle = {Proc.\ ACM SIGKDD International Conference on Knowledge Discovery \& Data Mining (KDD)},
  pages     = {3395--3404},
  year      = {2020},
}

@inproceedings{su2019omnianomaly,
  author    = {Su, Ya and Zhao, Youjian and Niu, Chenhao and Liu, Rong and Sun, Wei and Pei, Dan},
  title     = {Robust Anomaly Detection for Multivariate Time Series through Stochastic Recurrent Neural Network},
  booktitle = {Proc.\ ACM SIGKDD International Conference on Knowledge Discovery \& Data Mining (KDD)},
  pages     = {2828--2837},
  year      = {2019},
}

@article{tuli2022tranad,
  author    = {Tuli, Shreshth and Casale, Giuliano and Jennings, Nicholas R.},
  title     = {{TranAD}: Deep Transformer Networks for Anomaly Detection in Multivariate Time Series Data},
  journal   = {Proc.\ VLDB Endowment},
  volume    = {15},
  number    = {6},
  pages     = {1201--1214},
  year      = {2022},
}

@inproceedings{xu2022anomaly,
  author    = {Xu, Jiehui and Wu, Haixu and Wang, Jianmin and Long, Mingsheng},
  title     = {Anomaly Transformer: Time Series Anomaly Detection with Association Discrepancy},
  booktitle = {Proc.\ International Conference on Learning Representations (ICLR)},
  year      = {2022},
}

@inproceedings{deng2021gdn,
  author    = {Deng, Ailin and Hooi, Bryan},
  title     = {Graph Neural Network-Based Anomaly Detection in Multivariate Time Series},
  booktitle = {Proc.\ AAAI Conference on Artificial Intelligence},
  volume    = {35},
  number    = {5},
  pages     = {4027--4035},
  year      = {2021},
}

@inproceedings{breunig2000lof,
  author    = {Breunig, Markus M. and Kriegel, Hans-Peter and Ng, Raymond T. and Sander, J{\"o}rg},
  title     = {{LOF}: Identifying Density-Based Local Outliers},
  booktitle = {Proc.\ ACM SIGMOD International Conference on Management of Data},
  pages     = {93--104},
  year      = {2000},
}

@inproceedings{liu2008isolation,
  author    = {Liu, Fei Tony and Ting, Kai Ming and Zhou, Zhi-Hua},
  title     = {Isolation Forest},
  booktitle = {Proc.\ IEEE International Conference on Data Mining (ICDM)},
  pages     = {413--422},
  year      = {2008},
}

@article{ma2024mpformer,
  author    = {Ma, Shenhui and Nie, Jiahao and Guan, Siwei and He, Zhiwei and Gao, Mingyu},
  title     = {{MPFormer}: Multipatch Transformer for Multivariate Time-Series Anomaly Detection with Contrastive Learning},
  journal   = {IEEE Internet of Things Journal},
  volume    = {11},
  number    = {23},
  pages     = {38221--38237},
  year      = {2024},
}

@inproceedings{khosla2020supervised,
  author    = {Khosla, Prannay and Teterwak, Piotr and Wang, Chen and Sarna, Aaron and Tian, Yonglong and Isola, Phillip and Maschinot, Aaron and Liu, Ce and Krishnan, Dilip},
  title     = {Supervised Contrastive Learning},
  booktitle = {Advances in Neural Information Processing Systems (NeurIPS)},
  volume    = {33},
  pages     = {18661--18673},
  year      = {2020},
}

@inproceedings{radford2021clip,
  author    = {Radford, Alec and Kim, Jong Wook and Hallacy, Chris and Ramesh, Aditya and Goh, Gabriel and Agarwal, Sandhini and Sastry, Girish and Askell, Amanda and Mishkin, Pamela and Clark, Jack and Krueger, Gretchen and Sutskever, Ilya},
  title     = {Learning Transferable Visual Models from Natural Language Supervision},
  booktitle = {Proc.\ International Conference on Machine Learning (ICML)},
  pages     = {8748--8763},
  year      = {2021},
}

@inproceedings{kim2025carots,
  author        = {Kim, HyunGi and Mok, Jisoo and Lee, Dongjun and Lew, Jaihyun and Kim, Sungjae and Yoon, Sungroh},
  title         = {{CAROTS}: Causality-Aware Contrastive Learning for Robust Multivariate Time-Series Anomaly Detection},
  booktitle     = {Proc.\ International Conference on Machine Learning (ICML)},
  year          = {2025},
  archivePrefix = {arXiv},
  eprint        = {2506.03964},
}

@article{sanh2019distilbert,
  author    = {Sanh, Victor and Debut, Lysandre and Chaumond, Julien and Wolf, Thomas},
  title     = {{DistilBERT}, a Distilled Version of {BERT}: Smaller, Faster, Cheaper and Lighter},
  journal   = {arXiv preprint arXiv:1910.01108},
  year      = {2019},
  note      = {EMC\textsuperscript{2} Workshop at NeurIPS},
}

@inproceedings{vaswani2017attention,
  author    = {Vaswani, Ashish and Shazeer, Noam and Parmar, Niki and Uszkoreit, Jakob and Jones, Llion and Gomez, Aidan N. and Kaiser, {\L}ukasz and Polosukhin, Illia},
  title     = {Attention Is All You Need},
  booktitle = {Advances in Neural Information Processing Systems (NeurIPS)},
  volume    = {30},
  year      = {2017},
}

@inproceedings{loshchilov2019adamw,
  author    = {Loshchilov, Ilya and Hutter, Frank},
  title     = {Decoupled Weight Decay Regularization},
  booktitle = {Proc.\ International Conference on Learning Representations (ICLR)},
  year      = {2019},
}

@inproceedings{yang2023dcdetector,
  author    = {Yang, Yiyuan and Zhang, Chaoli and Zhou, Tian and Wen, Qingsong and Sun, Liang},
  title     = {{DCdetector}: Dual Attention Contrastive Representation Learning for Time Series Anomaly Detection},
  booktitle = {Proceedings of the 29th ACM SIGKDD Conference on Knowledge Discovery and Data Mining (KDD '23)},
  pages     = {3033--3045},
  year      = {2023},
  publisher = {ACM},
  doi       = {10.1145/3580305.3599295}
}

@inproceedings{wu2023timesnet,
  author    = {Wu, Haixu and Hu, Tengge and Liu, Yong and Zhou, Hang and Wang, Jianmin and Long, Mingsheng},
  title     = {{TimesNet}: Temporal 2D-Variation Modeling for General Time Series Analysis},
  booktitle = {The Eleventh International Conference on Learning Representations (ICLR)},
  year      = {2023}
}

@inproceedings{luo2024moderntcn,
  author    = {Luo, Donghao and Wang, Xue},
  title     = {{ModernTCN}: A Modern Pure Convolution Structure for General Time Series Analysis},
  booktitle = {The Twelfth International Conference on Learning Representations (ICLR)},
  year      = {2024},
  note      = {Spotlight}
}

@inproceedings{song2023memto,
  author    = {Song, Junho and Kim, Keonwoo and Oh, Jeonglyul and Cho, Sungzoon},
  title     = {{MEMTO}: Memory-guided Transformer for Multivariate Time Series Anomaly Detection},
  booktitle = {Advances in Neural Information Processing Systems (NeurIPS)},
  volume    = {36},
  year      = {2023}
}

@inproceedings{wang2023d3r,
  author    = {Wang, Chengsen and Zhuang, Zirui and Qi, Qi and Wang, Jingyu and Wang, Xingyu and Sun, Haifeng and Liao, Jianxin},
  title     = {Drift doesn't Matter: Dynamic Decomposition with Diffusion Reconstruction for Unstable Multivariate Time Series Anomaly Detection},
  booktitle = {Advances in Neural Information Processing Systems (NeurIPS)},
  volume    = {36},
  year      = {2023}
}

@article{cohen2025toto,
  author    = {Cohen, Ben and Khwaja, Emaad and Doubli, Youssef and Lemaachi, Salahidine and Lettieri, Chris and Masson, Charles and Miccinilli, Hugo and Ram{\'{e}}, Elise and Ren, Qiqi and Rostamizadeh, Afshin and Ogier du Terrail, Jean and Toon, Anna-Monica and Wang, Kan and Xie, Stephan and Xu, Zongzhe and Zhukova, Viktoriya and Asker, David and Talwalkar, Ameet and Abou-Amal, Othmane},
  title     = {This Time is Different: An Observability Perspective on Time Series Foundation Models},
  journal   = {arXiv preprint arXiv:2505.14766},
  year      = {2025}
}

@article{feofanov2025mantis,
  author    = {Feofanov, Vasilii and Wen, Songkang and Alonso, Marius and Ilbert, Romain and Guo, Hongbo and Tiomoko, Malik and Pan, Lujia and Zhang, Jianfeng and Redko, Ievgen},
  title     = {{Mantis}: Lightweight Calibrated Foundation Model for User-Friendly Time Series Classification},
  journal   = {arXiv preprint arXiv:2502.15637},
  year      = {2025}
}

@inproceedings{nie2023patchtst,
  author    = {Nie, Yuqi and Nguyen, Nam H. and Sinthong, Phanwadee and Kalagnanam, Jayant},
  title     = {A Time Series is Worth 64 Words: Long-term Forecasting with Transformers},
  booktitle = {The Eleventh International Conference on Learning Representations (ICLR)},
  year      = {2023}
}

@inproceedings{liu2024itransformer,
  author    = {Liu, Yong and Hu, Tengge and Zhang, Haoran and Wu, Haixu and Wang, Shiyu and Ma, Lintao and Long, Mingsheng},
  title     = {{iTransformer}: Inverted Transformers Are Effective for Time Series Forecasting},
  booktitle = {The Twelfth International Conference on Learning Representations (ICLR)},
  year      = {2024},
  note      = {Spotlight}
}

@inproceedings{zhou2023gpt4ts,
  author    = {Zhou, Tian and Niu, Peisong and Wang, Xue and Sun, Liang and Jin, Rong},
  title     = {One Fits All: Power General Time Series Analysis by Pretrained {LM}},
  booktitle = {Advances in Neural Information Processing Systems (NeurIPS)},
  volume    = {36},
  year      = {2023},
  note      = {Spotlight}
}

\end{document}